\documentclass[letterpaper, 10 pt, conference]{ieeeconf}  

\IEEEoverridecommandlockouts                              

\usepackage{mathptmx} 
\usepackage{times} 
\usepackage{amsmath} 
\usepackage{amssymb}  
\usepackage{tabularx} 
\usepackage{graphicx}
\usepackage{comment}
\usepackage{mathtools}
\usepackage{upgreek}
\usepackage{gensymb}
\usepackage{tabularx}   
\usepackage{array}
\usepackage{booktabs}
\newcolumntype{Y}{>{\raggedright\arraybackslash}X}
\usepackage[noadjust]{cite}
\usepackage{placeins}
\usepackage{xcolor}
\usepackage{url}
\usepackage{mwe}
\usepackage{balance}
\usepackage{multirow}
\usepackage{amsmath}
\usepackage{booktabs}
\DeclareMathOperator*{\argmin}{arg\,min}
\usepackage[font=footnotesize,labelfont=bf]{caption}

\usepackage{dblfloatfix}
\newcolumntype{L}[1]{>{\raggedright\let\newline\\\arraybackslash\hspace{0pt}}m{#1}}
\newcolumntype{C}[1]{>{\centering\let\newline\\\arraybackslash\hspace{0pt}}m{#1}}
\newcolumntype{R}[1]{>{\raggedleft\let\newline\\\arraybackslash\hspace{0pt}}m{#1}}

\newcommand{\FK}{\text{\textit{FK}}}
\newcommand{\IK}{\text{\textit{IK}}}

\usepackage{mathtools}
\usepackage{wasysym}
\usepackage{bm}
\usepackage{bbm}
\usepackage{array}

\usepackage{algpseudocode}
\usepackage[linesnumbered,ruled,vlined,noend]{algorithm2e}
\SetKwRepeat{Do}{do}{while}

\DeclarePairedDelimiter\abs{\lvert}{\rvert}%
\DeclarePairedDelimiter\norm{\lVert}{\rVert}%

\makeatletter
\let\oldabs\abs
\def\abs{\@ifstar{\oldabs}{\oldabs*}}
\let\oldnorm\norm
\def\norm{\@ifstar{\oldnorm}{\oldnorm*}}
\makeatother

\title{\LARGE \bf
Bi-manual Stabilization of the Cervical Spine for Safe Physical Human-Robot Interaction in Rolling Maneuvers
}

\author{Elizabeth Peiros$^{1}$, Moira Bohley$^{1}$, Lucas Yager$^2$, Derek Chen$^{1}$, Rebecca Fellbaum$^{1}$, \\ Chris D'Ambrosia$^3$, Ananya Rao$^{1}$, Xiao Liang$^{1}$, Michael C. Yip$^1$ \IEEEmembership{Senior Member, IEEE}
\thanks{$^1$ Elizabeth, Moira, Derek, Rebecca, and M.C. Yip are with the Electrical and Computer Engineering Department, University of California, San Diego, La Jolla, CA 92093 USA. {\tt\footnotesize\{epeiros, dbohley, dec004, rfellbaum, anr126, x5liang, yip\}@ucsd.edu}}%
\thanks{$^2$ Lucas Yager is with the Mechanical and Aerospace Engineering Department, University of California, San Diego, La Jolla, CA 92093 USA.{\tt\footnotesize \{lyager\}@ucsd.edu}}%
\thanks{$^3$ Christopher D'Ambrosia with Columbia University Irving Medical Center, National Outdoor Leadership School Wilderness Medicine Institute. {\tt\footnotesize \{cd2859\}@cmc.columbia.edu}}%
}

\begin{document}

\maketitle
\thispagestyle{empty}
\pagestyle{empty}


\begin{abstract}
The neck, specifically the cervical spine, is a highly fragile and mobile part of the body with a high risk of catastrophic injury. Whether someone is injured on a battlefield, in a natural disaster, or in an athletic event, great care is always taken with the cervical spine when transporting, rolling, or moving the patient. In this work, we present an algorithm and adaptations for bi-manual robots to stabilize the cervical spine during rolling maneuvers.
We investigate techniques to maximize grip strength within safety bounds and present a control approach for proper neck rotation.
We found that with our approach, a bi-manual robot is capable of achieving human-level performance in maintaining target rotational alignments. The axial alignment for the robot with prediction was within .21\degree of human novices.
\end{abstract}

\section{Introduction}
\vspace{-1mm} 


The neck, specifically the cervical spine, is a highly fragile and mobile part of the body with a high risk of catastrophic injury in major events. It is a focal point when there is ever consideration for spinal injury or head injury: car crashes, falls, collisions in sports, explosions. In field medicine, it is a major consideration during patient transport and maneuvering. Robots have been used in practice to lift patients \cite{madan2026tactic,mukai2010development} or reposition limbs \cite{peiros2023finding}, but have not yet been used for a delicate and important role such as cervical spine repositioning and stabilization. 

Patient transfers and extractions are common across multiple domains. Military corpsmen and wilderness medicine experts are trained to perform these tasks with minimal personnel and equipment, while nurses, EMTs, and other first responders are trained to use equipment and work with large teams to perform these transfers \cite{bergman2020patient}. A common requirement across all transfers is a focus on the head and neck to stabilize injury, maintain an airway, and organize cables, pumps, or central lines during transfer~\cite{bergman2020patient}. 

In regard to stabilizing strategy, the rigid spinal collar, which is made of plastic and placed around the neck to restrict all cervical spine motion during transport, is likely what comes to mind. However, despite their prevalence, these are not considered best practice. Current practice has moved away from rigid collars because research shows that they can increase intracranial pressure and blood circulation \cite{davies1996effect, hunt2001effects, swartz2018prehospital, liao2018motion}. Evidence also suggests that rigid collars can increase the risk of hypoxia by reducing airflow to the lungs \cite{ala2016cervical}.

A rigid collar is nonetheless used when personnel and time are limited, as military corpsmen experience when evacuating a wounded warrior from active combat. In field medicine, specifically wilderness medicine, the recommendation when solo and lacking in supplies is to use an article of clothing as a soft collar \cite{porter2019improvised}. While this is a solution, the ideal case is to designate a person to head stabilization when performing any rolling maneuver. A perfect robotic teammate in this scenario could perform the desired neck stabilization to avoid using collars entirely. The robot would use a bimanual grasp of the head around the ears, as humans would, circumventing concerns about compressing critical anatomy. 

\begin{figure}
\centering
    \includegraphics[trim=18cm 5cm 19cm 11cm, clip, width=0.99\linewidth]{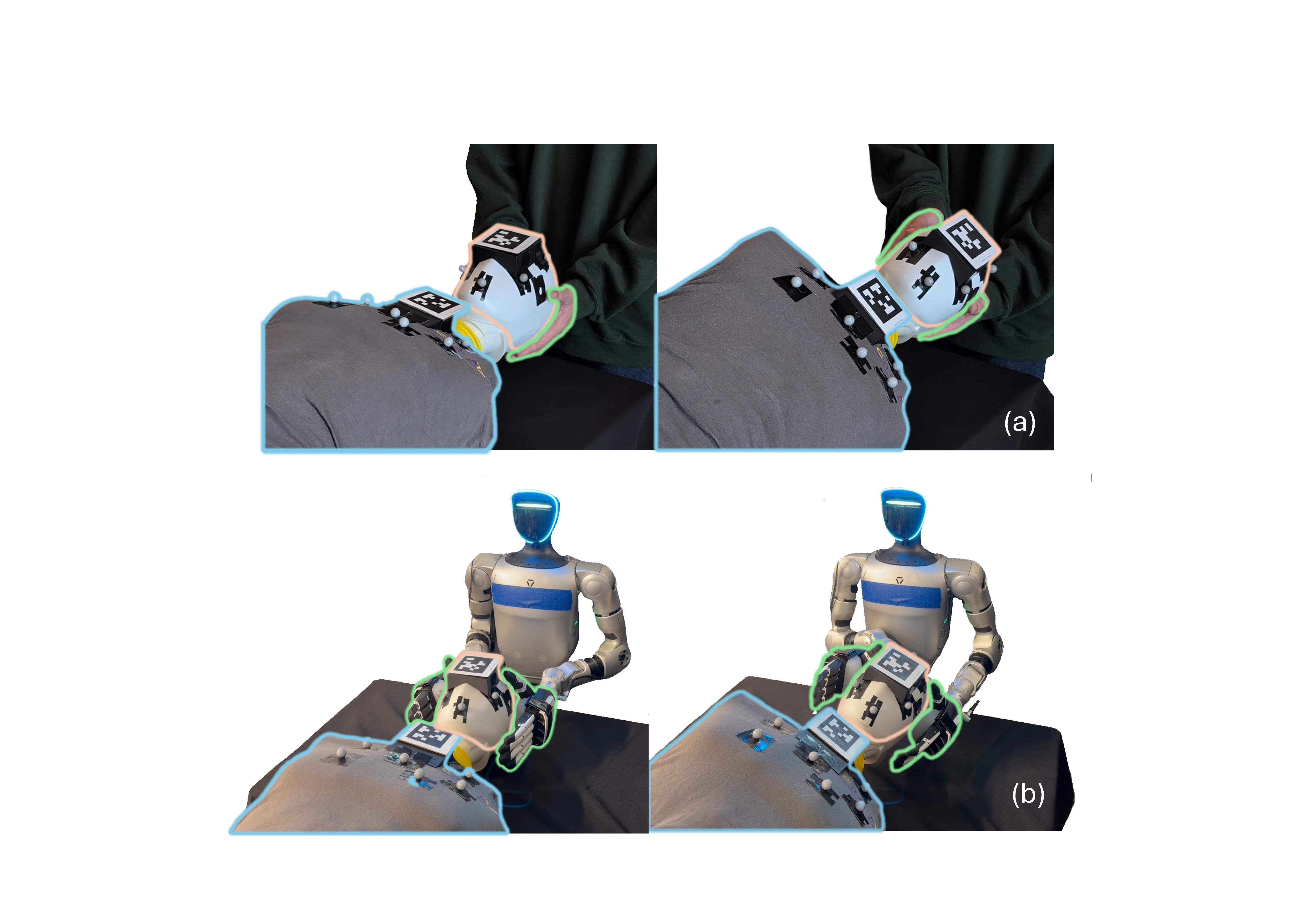}
    \vspace{-8mm}
    \caption{(a) Human performing neck stabilization task; (b) bi-manual robot platform executing neck stabilization task}
    \label{fig:title-image}
    \vspace{-5mm}
\end{figure}

In this work, we present an algorithm that allows a bi-manual robot to act as a cervical spine stabilizer during rolling maneuvers. Specifically, we:

\begin{itemize}
    \item Developed an algorithm that uses available body tracking methods to define trajectories for neck stabilization, maintaining a straight and uncompressed neck with ears over shoulders
    \item Created deformable hand pads optimized to maximize the ratio of input pressure to output grasp force to both reduce work done by the robot to maintain grip and meet biomechanical safety constraints.
    \item Designed a testing phantom that measured relevant forces and angular displacements to evaluate the capabilities of the robot relative to humans.
\end{itemize}

\section{Background}
\label{humanModel: Neck}



\subsubsection{Cervical Spine Anatomy and Biomechanics}

\begin{figure}
    \centering
    \includegraphics[trim=2cm 21cm 10cm 20cm, clip, width=0.99\linewidth]{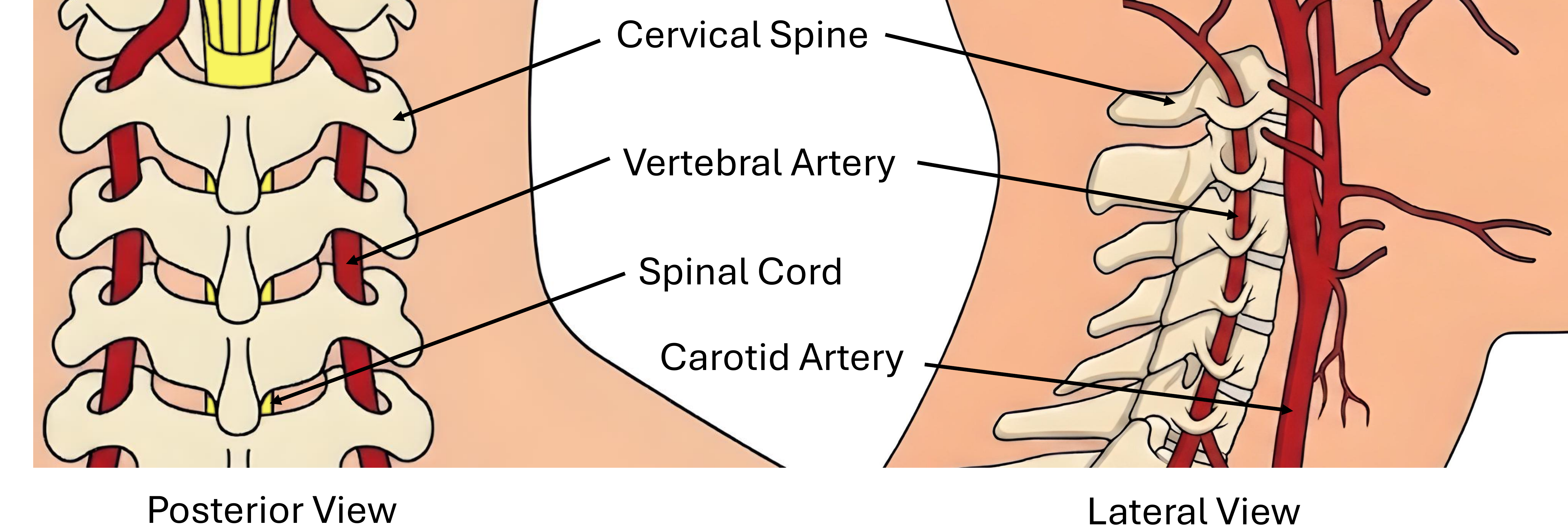}
    \caption{The neck contains not only the major nerve bundle that makes up the spinal cord responsible for all nerve function below, but it also has major arteries that run to either side and provide the blood supply to the brain.}
    \label{fig: neck_anatomy}
    \vspace{-6mm}
\end{figure}

The C-spine is the portion of the spine that starts at the base of the skull and extends to the shoulders. It consists of 7 vertebrae, the last of which can be found by palpating for the nodule at the base of the back of the neck. The neck contains not only the major nerve bundle that makes up the spinal cord, responsible for all nerve function below, but also major arteries that run to either side, providing the brain's blood supply. The major anatomical features can be seen in Fig \ref{fig: neck_anatomy}. The range of motion of the neck in each degree of freedom of rotation can be seen in Table \ref{tab: neck_joint_limits}.

\begin{table}[t]
\vspace{2mm}
\centering
\caption{Human model parameters: Neck Joint Limits}
\label{tab: neck_joint_limits}
\begin{tabular}{cccc}
    \toprule
    \multicolumn{3}{c}{\textbf{Cervical Spine: Range of Motion}} & \textbf{Max Allowable Dev}\\
    \midrule
    Flexion               & $54.4\degree$ &                  &  $10\degree$  \\
    Extension             & $93.2\degree$ & total $147.6\degree$ & $10\degree$  \\
    Lateral Bending (L)   & $40\degree$     &                  & $10\degree$  \\
    Lateral Bending (R)   & $40\degree$     &                  & $10\degree$  \\
    Left Axial Rotation   & $75\degree$   &                  & $10\degree$  \\
    Right Axial Rotation  & $76\degree$   & total $151\degree$ & $10\degree$  \\
    \bottomrule
\end{tabular}\\
\vspace{1mm}
\footnotesize\textit{Ranges for flexion and rotation are taken from \cite{bennett1963range}, which focused on women aged 18--24. These measurements are taken using a goniometer from a neutral starting position. Lateral bending values were taken from \cite{gilman1979measurement}}
\vspace{-2mm}
\end{table}

\subsubsection{Cervical Spine Injury Prevention and Stabilization Techniques}

For injury prevention during intervention, the main considerations are avoiding C-spine damage by not applying external load to the neck and avoiding head movement relative to the torso and shoulders. Medical studies have shown that it is often necessary to realign the neck to a neutral position to avoid compressing the carotid arteries or nerves of the neck, which can lead to paralysis or brain damage\cite{davies1996effect, swartz2009cervical}. During this realignment action, one must avoid introducing any compression, distraction, or lateral forces to the c-spine, and prevent rotational motions. Not adhering to these safety measures could mean misalignment of damaged vertebrae. With our wilderness medicine/clinical collaborator we have defined the allowable rotation based on the "ears over shoulders" metric described to humans executing such a task. The exact allowable deviations are found in table \ref{tab: neck_joint_limits}. At this time force goals are to aim to meet the same minimum forces human stabilizers would be able to achieve.

\subsubsection{Safe Grasping Forces}
\label{sec: bioGrip}

When providing enough grip to stabilize the head, it is important to regulate pressure and contact area to avoid damaging the underlying vessels or, more importantly, applying extraneous pressure to areas that could block the blood supply, such as the temples. While there is not an exact pressure metric for grasping research has examined the threshold for impact bruising; current understanding suggests that pressure or energy transfer over an area drives this type of injury. In one study they found that injury occurs after $1.3 \; MPa$ and total energy transfer is more than $87 kJ/m^2$ \cite{8204347}. These values were derived from porcine data using equations \ref{eq:peak_pressure} and \ref{eq:energy_transfer} where $P_{peak}$ is the peak pressure measured from the impact, $F_{peak}$ is the peak force, $A$ is the contact surface area, $U_{total}$ is the total energy displaced to the tissue on impact and $\delta_{max}$ is the maximum displacement of the tissue:
\vspace{-2mm}
\begin{equation}
\vspace{-3mm}
\label{eq:peak_pressure}
    P_{peak} = F_{peak}/A
\end{equation}

\begin{equation}
\vspace{-1mm}
\label{eq:energy_transfer}
    U_{total} = \frac{1}{A}\int_{0}^{\delta_{max}} F \,d\delta
\end{equation}

The injury mechanism is described similarly to plastic deformation, where $U_{net}$ is the energy loss due to plastic deformation during impact. Plastic deformation is the region in which deformable objects permanently deform by absorbing energy beyond their ability to restore that energy.

\begin{equation}
\label{eq:energy_net}
    U_{net} = \frac{1}{A} \left( \int_{0}^{\delta_{max}} F \,d\delta-\int_{\delta_{max}}^{0} F \,d\delta \right)
\end{equation}

\section{Related Works}

\subsubsection{Rescue Robotic Platforms}
Numerous competitions for developing human rescue platforms have occurred in Europe and the US. ELROB, a competition held in Europe in 2014, yielded numerous robotic platforms designed to rescue casualties from conflict zones, including platforms with mechanisms like fishing reels for dragging and elevator lifts\cite{10018818,bruggemann2016search}. In the United States, platforms such as the Battlefield Extraction-Assistant Robot (BEAR)\cite{theobald2010mobile,watts2004tatrc,murphy2004robot}, ResQBot\cite{saputra2018resqbot}, ResQBot2.0\cite{saputra2021resqbot}, and HURCULES\cite{choi2019development} are designed to carry casualties using industrial manipulators and conveyor belts. These platforms are uniquely designed for casualty extraction in terms of strength and size. However, they are not designed for complex, fine, and constrained manipulation tasks, nor do they consider biomechanics or safety during strategic contact or operation.

\subsubsection{Human Manipulation with General Robotic Platforms}

Researchers have expanded the capabilities of generic manipulators to physically assist people during chair transfers and physical therapy delivery by leveraging augmented general-purpose platforms for compliant interactions. Those focused on rolling explicitly study the whole-body roll and ignore c-spine stabilization \cite{miyake2025deep,matsumura2025automated, madan2026tactic}.

\subsubsection{Bi-manual Constrained Robotics Tasks}

Bi-manual tasks are most commonly concentrated on tools and copying the visual motion of humans performing similar tasks. Many learn to interact with articulating objects through videos \cite{mandi2025dexmachina, kerr2024robot}, while others utilize simulation and RL to build semantic understanding of how to articulate hinges or lids \cite{zhu2026learning, sun2025vtao}. Other techniques which are followed in this paper use kinematic constraints\cite{liang2025towards, cohn2024constrained}. Some have also done bi-manual rotation of objects \cite{lertkultanon2018certified}. None have used these techniques with safety constraints, and most stick to single rigid bodies.

\section{Methods}

The overall method is broken down into sub-tasks which are completed in the following sequence.  

\subsubsection*{Step 1}
\label{Step 1} The head-mounted camera on the bimanual robotic platform locates two ArUco markers: one attached to the casualty head and one attached to the casualty torso. The robot records the location of the head marker relative to the robot's base frame.
\subsubsection*{Step 2}\label{Step 2}The robot runs a planning algorithm with the goal of moving its hands such that each hand is on either side of the head. The robot then executes the motion by moving the hands to those positions using $RRT^*$ with collision detection in joint space and a uses PD controller for end-effector position control for execution.
\subsubsection*{Step 3}\label{Step 3}The robot runs a planning algorithm with the goal of moving its hands such that each hand is on either side of the head. The robot then executes the motion by moving the hands to those positions using $RRT^*$ with collision detection in joint space and uses a PD controller for end-effector position control for execution.
\subsubsection*{Step 4}\label{Step 4}The dexterous hands move toward the head to apply a designated pressure/force to the sides of the head defined by optimizing for maximizing grasping strength relative to a minimal input pressure on the contact points. The appropriate grip force is used to maintain control of the head for stabilization motions.
\subsubsection*{Step 5}\label{Step 5}The robot identifies the head-torso misalignment, then aligns the neck by matching the head and torso orientations based on the angular difference between the two.
\subsubsection*{Step 6}\label{Step 6}The robot continuously tracks the torso and head orientations to maintain alignment when the torso is rolled without introducing undesired forces by taking advantage of a torso motion model.

This task breakdown draws on medical journals and lived experience from experts in the field discussed in section \ref{humanModel: Neck}.

\subsection{Pose Tracking and Coordinate Frames}
\label{sec:frames}

For visual tracking, ArUco markers\cite{garrido2014automatic} were chosen as the position-and-orientation tracking system. While other marker-based and markerless tracking systems exist, the ArUco markers provided a few key features: (1) they can be applied and oriented to match the clinical angles and vectors used in medical and
biomechanics work; (2) they integrate with the robotic pipeline using the
robot's head-mounted camera, which better translates to a system for field
use; (3) markerless tracking requires specific human face features for
registration, which are not present.

We write $^{A}T_{B} \in SE(3)$ for the transform mapping points from frame $B$
into frame $A$, so chains compose right-to-left and
$^{A}T_{B} = (^{B}T_{A})^{-1}$; $\mathrm{Log}: SO(3) \to \mathbb{R}^{3}$ is the
rotation-vector logarithm and $[\cdot]_{\times}$ its inverse. The frames are
listed in Table~\ref{tab:frames}. The robot approaches the casualty using
locomotion but stabilizes from a planted stance, so the base frame $B$ is
stationary over an episode and serves as the inertial frame; it is upright on
level ground, making $\hat{\mathbf{z}}_{B}$ gravity-aligned.

\begin{table}[t]
\vspace{2mm}
\centering
\caption{Coordinate frames.}
\label{tab:frames}
\small
\setlength{\tabcolsep}{3pt}
\renewcommand{\arraystretch}{1.15}
\begin{tabularx}{\columnwidth}{@{}l Y Y >{\raggedright\arraybackslash}p{1.6cm}@{}}
\toprule
\textbf{Frame} & \textbf{Attached to} & \textbf{Obtained from} &
\textbf{Status} \\
\midrule
$B$ & Robot base & --- & Assumed Fixed \\
$C$ & Camera & Robot FK & Constant \\
$M_{h}$, $M_{t}$ & ArUco markers on head, torso & Marker detection & Measured \\
$H$, $Tr$ & Head, torso biomechanics frames & Fixed offsets $^{M_h}T_{H}$, $^{M_t}T_{Tr}$ & Calibrated constants \\
$L$, $R$ & Left, right end-effectors & Forward kinematics $\mathrm{FK}(\mathbf{q})$ & Measured \\
$F_{0}$ & Torso pose at episode start & Snapshot of $^{B}T_{Tr}$ at $k=0$ & Constant per episode \\
\bottomrule
\end{tabularx}
\vspace{-6mm}
\end{table}

One marker is affixed to the head and one to the torso, each carrying a fixed
calibrated offset into its biomechanics frame, since the native marker axes do
not coincide with the clinical rotation directions of
Table~\ref{tab: neck_joint_limits}. The head and torso poses in the base frame are
\begin{equation}
^{B}T_{H} = {}^{B}T_{C}\; ^{C}T_{M_h}\; ^{M_h}T_{H},
\qquad
^{B}T_{Tr} = {}^{B}T_{C}\; ^{C}T_{M_t}\; ^{M_t}T_{Tr},
\end{equation}
with $^{B}T_{C}$ the calibrated camera-to-base transform and $^{M_h}T_{H}$,
$^{M_t}T_{Tr}$ the marker-to-body offsets measured once during setup. The head pose relative to the torso, described in Step 1, follows by composition,
\vspace{-2mm}
\begin{equation}
^{Tr}T_{H} = \big(^{B}T_{Tr}\big)^{-1}\, ^{B}T_{H},
\end{equation}
whose rotational part $^{Tr}R_{H}$ is decomposed as an intrinsic $z$--$y$--$x$
Euler sequence about the torso axes, mapping to [axial rotation,
flexion/extension, lateral bending]. These are the clinically
relevant angles limited in Table~\ref{tab: neck_joint_limits}. 

The episode frame $F_{0}$ is the torso pose at the first step of a rolling
episode,
\begin{equation}\label{eq:episodeframe}
^{0}T_{Tr}(k) = \big(^{B}T_{Tr}(0)\big)^{-1}\, ^{B}T_{Tr}(k),
\end{equation}
used in Section~\ref{sec:gp} so the learned prior is independent of where the
robot stands relative to the casualty. Since $B$ is stationary, the anchor
$^{B}T_{Tr}(0)$ is constant and $^{B}T_{Tr}(k)$ is recovered without odometry.

\subsection{Move to Head}

For Step 2 and 3, sample-based planning was used to plan motion for both arms to the head's location to grasp it safely.

\subsubsection{Planning Trajectory to the Head}

Once the head pose is known in the robot's base frame, the starting and goal pose of both arms' end effectors can be determined. The goal pose is one robotic hand to the left and one to the right of the head, by the ears, with palms facing inward. The goal locations are defined in the head frame as the geometric center $\pm (\frac{BB_H}{2} + 4$) where $BB_H$ is the approx head width measured in cm. The goal orientation is defined in the world frame as facing toward each other, with the line of action between the hands parallel to the floor plane ($\cdot \hat{z} = 0$). In later steps, this will force any neck deviations in the axial rotation direction to realign with hand closure. The right arm's goal pose is assumed to have its origin at the point farthest to the right, and the left arm's at the point farthest to the left. 

The collision objects considered in planning are the floor (with a constraint on hand position relative to z-height), the hands with each other, the head, and the torso, all described by bounding primitives. The hands and torso have bounding boxes, while the head has bounding spheres. To define the initial path, $RRT^*$ \cite{karaman2011sampling} is used with a step size of .3 rad, a rewire radius of .6 rads and a maximum of 9000 iterations. With these considerations, a linear Cartesian-space trajectory is formed using Smoothstep \cite{ebert2003texturing} to create a smooth series of waypoints; then LERP and SLERP \cite{shoemake1985animating} are used to interpolate the path back into nodes.

\subsubsection{Gravity Compensation for Hands}

Gravity compensation was added to account for the attached hands and arms' overall mass. Inverse dynamics was used to find the feedforward motor torques required to compensate using the Recursive Newton-Euler Algorithm\cite{luh1980line} during non-contact motion when the arms remain open chains.

\subsection{Grasping Head}
\subsubsection{Executing Grasp}
\label{sec: head grasp}
Once the hands reach their goal poses on either side of the head, they move steadily inward toward the head along the head frame's $\pm \hat{h}_y$ as described in Step 4. Once contact is made, the wrench force is measured, and the hands slowly squeeze inward until the threshold for "good grip" is reached, determined by calibrating the pressure-to-grasp-force ratios. Trajectories are executed using the same formulation shown in equation \ref{eq:qp}. This process uses the wrench force to set the desired position, $\boldsymbol{p}_y \in \mathbb{R}^2$, used to maintain grip, making it generalizable to different head sizes.

\subsubsection{Grip Strength}
As discussed in section \ref{sec: bioGrip}, the pressure on each contact regions should not exceed 1.3MPa. Grasp force balances providing maximal grip force, defined as resistance to disturbance, with maintaining pressure below the threshold. To meet these conflicting demands, an optimization framework was derived to compare bare dexterous hands with the use of compliant pads. This optimization models the pads as deformable spheres is pushing into a rigid flat wall.
\vspace{-3mm}
\begin{equation}
\label{eq: material_opt}
\begin{aligned}
\max_{d \in \mathbb{R}^n} \quad & \frac{f_r(f_y, A_c)}{P(f_y,A_c)} \\
\text{subject to} \quad & \frac{f_y}{A_c} \leq 1.3MPa \\
& A_c = \pi a^2 = \pi\left(\frac{3f_yR(1-\nu^2)}{4E}\right)^{2/3}
\end{aligned}
\end{equation}
Where $d$ is the durometer of the silicone material, $f_r$ is the peak resistance force before loss of grip, $f_y$ is the force pressing into the surface (into the head), $A_c$ is the area of the contact patch, $P$ is the pressure, $R$ is undeformed radius of the sphere, $E$ is Young's modulus of the material, and $\nu$ is Poisson's ratio.

\subsection{Bi-manual Stabilization and Control}\label{sec:stable_control}





\subsubsection{Planning and Pose Tracking}
With the head in the hands, the robot switches to the bi-manual stabilization
algorithm, maintains alignment of the head with the
torso/shoulders as described in Steps 5 and 6. The markers, instead of tracking solely head location, are now
used to measure the head angle relative to the torso to define real-time
trajectories that reorient and move the head into alignment.

When the grasp closes at step $k_{0}$, each hand becomes rigidly attached to
the head, fixing the constant offsets
\begin{equation}
\label{eq:graspoffsets}
^{H}T_{L} = \big(^{B}T_{H}(k_{0})\big)^{-1}\, ^{B}T_{L}(k_{0}),
\qquad
^{H}T_{R} = \big(^{B}T_{H}(k_{0})\big)^{-1}\, ^{B}T_{R}(k_{0}).
\end{equation}
The desired head pose is specified relative to the torso as
$^{Tr}T_{H}^{*} = \big(R^{*},\, ^{Tr}\mathbf{p}_{H}(k_{0})\big)$, where $R^{*}$
is the neutral ``ears over shoulders'' orientation of Section~\ref{humanModel: Neck}
and the translation is held at its value at grasp, so the head is reoriented
without introducing compression or distraction of the c-spine. Pushing this
through the grasp offsets gives the hand targets in the base frame,
\begin{equation}
\label{eq:handtargets}
^{B}T_{L}^{*} = {}^{B}T_{Tr}\; ^{Tr}T_{H}^{*}\; ^{H}T_{L},
\qquad
^{B}T_{R}^{*} = {}^{B}T_{Tr}\; ^{Tr}T_{H}^{*}\; ^{H}T_{R},
\end{equation}
from which $\mathbf{p}^{*}_{L}, R^{*}_{L}$ and $\mathbf{p}^{*}_{R}, R^{*}_{R}$
are read off. Once the goal pose is reached it becomes the saved pose, and
Eq.~(\ref{eq:handtargets}) is re-evaluated at each step from the current torso
measurement.

The angular difference is fed through a median filter, then an exponential moving average (EMA) filter, and finally a slew-rate limiter. The median filter catches large unexpected spikes in the input stream if the computer vision misreads the marker as far from its actual position, while the EMA filter
smooths out jitter, which can be quite common in tracking ArUco markers. The slew-rate limiter then caps the robot motion, preventing extreme motions that would violate the safety and actuator constraints of the robot arms.

\subsubsection{Trajectory Execution Real Time in Joint Space}
This execution follows a similar paradigm to the dual-arm motion described
above; however, it removes the collision constraints previously included in
planning and adds a force-regulator term at each step to maintain a strong
grasp.
\begin{equation}\small
\label{eq:qp}
\begin{split}
\mathbf{q}^{*}_{k} = \operatorname*{arg\,min}_{\mathbf{q} \in \mathbb{R}^{14}}
\; &\alpha_{1}\big\|\mathbf{e}_{p}(\mathbf{q};\, T^{k}_{L}, T^{k}_{R})\big\|^{2}
+ \alpha_{2}\big\|\mathbf{e}_{R}(\mathbf{q};\, T^{k}_{L}, T^{k}_{R})\big\|^{2}\\
&
+ \alpha_{3}\big\|\mathbf{q} - \mathbf{q}_{k-1}\big\|^{2}\\[4pt]
\end{split}
\end{equation}
where $\mathbf{q} \in \mathbb{R}^{14}$ are the joint angles of both arms and
$[\alpha_{1}=50, \alpha_{2}=25, \alpha_{3}=0.1]$ are hyper-parameters that we tune. The first two terms emphasize error
minimization, the third minimizes the joint angles to incentivize moving away
from singular configurations, and the final term minimizes step size to
maintain smoothness in the trajectory.

Both $\mathrm{FK}(\mathbf{q})$ and the targets of
Eq.~(\ref{eq:handtargets}) are expressed in the base frame $B$, so the stacked
task errors below are well posed in that frame:
\begin{equation}
\label{eq:errors}
\begin{aligned}
\mathbf{e}_{p}(\mathbf{q}) &=
\begin{bmatrix}
\mathrm{FK}(\mathbf{q}_{L}).\mathbf{p} - \mathbf{p}^{*}_{L}\\
\mathrm{FK}(\mathbf{q}_{R}).\mathbf{p} - \mathbf{p}^{*}_{R}
\end{bmatrix} \in \mathbb{R}^{6},\\[6pt]
\mathbf{e}_{R}(\mathbf{q}) &=
\begin{bmatrix}
\mathrm{Log}\big(R_{L}(\mathbf{q}_{L}) R_{L}^{*\top}\big)\\
\mathrm{Log}\big(R_{R}(\mathbf{q}_{R}) R_{R}^{*\top}\big)
\end{bmatrix} \in \mathbb{R}^{6}.
\end{aligned}
\end{equation}
Here $\mathrm{FK}(\mathbf{q}_{L}).\mathbf{p}$ and
$R_{L}(\mathbf{q}_{L})$ denote the end-effector position and orientation
computed via forward kinematics, and $\mathrm{Log}(\cdot)$ maps the spatial
orientation error in $SO(3)$ to its corresponding angular-error vector in
$\mathbb{R}^{3}$.

\subsubsection{Prediction in Torso Motion}
\label{sec:gp}
As is inherent in all systems, there is a lag between the reading of the torso
pose and the actions of the arms to keep the head alignment. A Gaussian Process
(GP) based prediction method is developed to reduce lag by learning the pattern
during the rolling maneuver.

\begin{figure}
\vspace{2mm}
    \centering
    \includegraphics[trim=4cm 21cm 4cm 8cm, clip, width=0.99\linewidth]{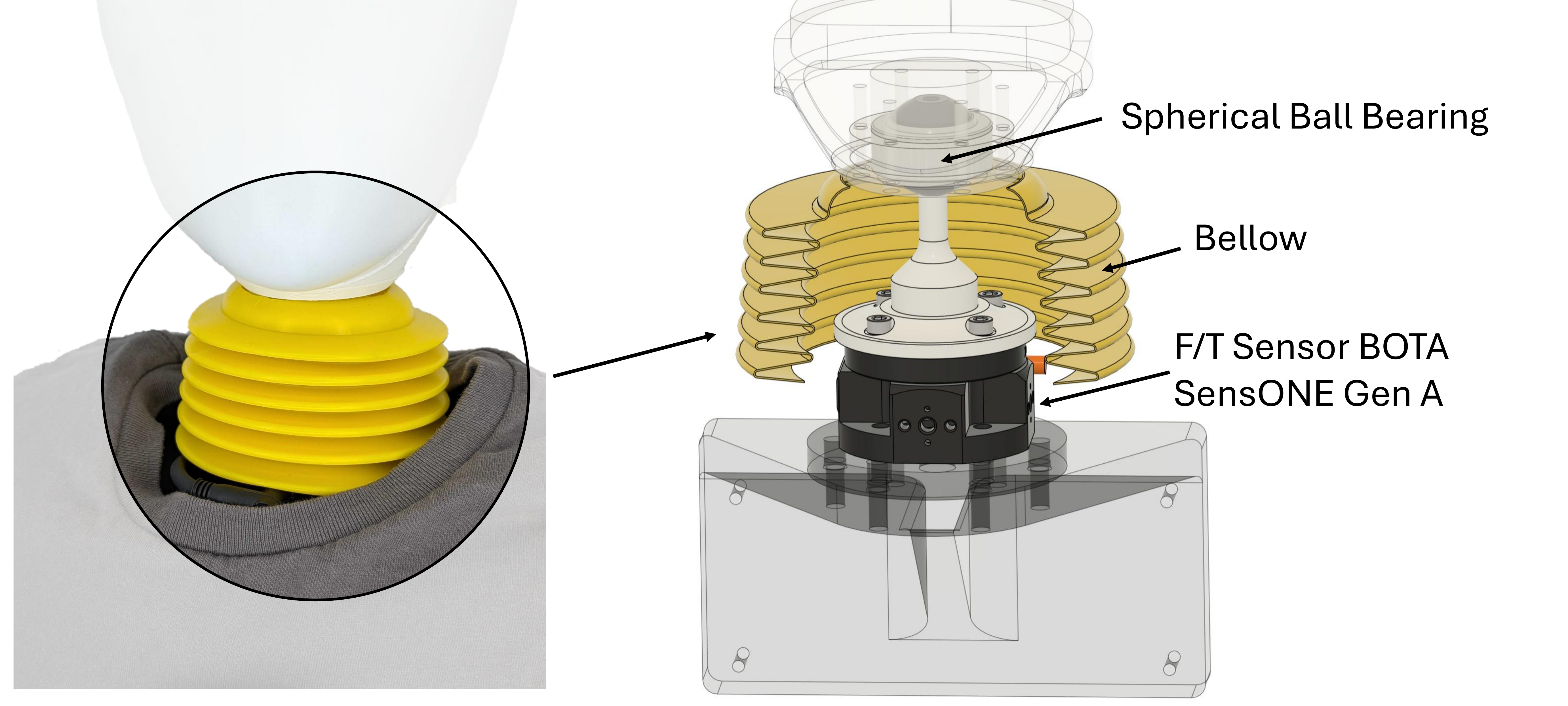}
    \caption{Phantom used for testing which features a F/T load cell and bio-mechanically representative neck joint. }
    \label{fig: testing_man}
    \vspace{-6mm}
\end{figure}

Let a torso state vector at time $k$ be defined as
\begin{equation}
\label{eq:gpstate}
\mathbf{x}_{k} = \big[\, \mathbf{p}_{k},\; \mathbf{r}^{(1)}_{k},\;
\mathbf{r}^{(2)}_{k},\; \mathbf{v}_{k},\; \boldsymbol{\omega}_{k} \,\big]
\in \mathbb{R}^{15},
\end{equation}
where $\mathbf{p}_{k}$ is the position and
$\mathbf{r}^{(1)}_{k}, \mathbf{r}^{(2)}_{k}$ the first two columns of the
orientation matrix, both taken from $^{0}T_{Tr}(k)$ of
Eq.~(\ref{eq:episodeframe}), and $\mathbf{v}_{k}, \boldsymbol{\omega}_{k}$ are
backward-difference linear and angular velocities in the body frame at $k$. The
regression target is the pose change over the next step, expressed in the body
frame at time $k$,
\begin{equation}
\label{eq:gptarget}
\mathbf{y}_{k} = \big[\, R_{k}^{\top}(\mathbf{p}_{k+1} - \mathbf{p}_{k}),\;
\mathrm{Log}\big(R_{k}^{\top} R_{k+1}\big) \,\big] \in \mathbb{R}^{6}.
\end{equation}
Each of the six target dimensions is modeled by an independent scalar Gaussian
process,
\begin{equation}
\label{eq:gp}
y = f(\mathbf{x}) + \varepsilon, \quad
f \sim \textit{GP}\big(0,\, k(\mathbf{x}, \mathbf{x}')\big), \quad
\varepsilon \sim \mathcal{N}(0, \sigma_{n}^{2}),
\end{equation}
with the anisotropic squared-exponential (ARD-RBF) kernel
$k(\mathbf{x}, \mathbf{x}') = \sigma_{f}^{2}
\exp\big(-\tfrac{1}{2}\sum_{d=1}^{15} (x_{d}-x'_{d})^{2}/\ell_{d}^{2}\big)$.
Each output dimension is fit as an exact GP on $200$ input--target pairs
subsampled from $40$ training episodes, with kernel hyperparameters set by
maximizing the marginal likelihood~\cite{seeger2004gaussian}.

At runtime, the GP mean
$\hat{\mathbf{y}} = [\hat{\mathbf{y}}_{p}, \hat{\mathbf{y}}_{r}]$ evaluated at
the current state augments the current measured pose
$(\mathbf{p}_{k}, R_{k})$ into the predicted next pose,
\begin{equation}
\label{eq:gppredict}
\hat{\mathbf{p}}_{k+1} = \mathbf{p}_{k} + R_{k}\,\hat{\mathbf{y}}_{p},
\qquad
\hat{R}_{k+1} = R_{k} \exp\big([\hat{\mathbf{y}}_{r}]_{\times}\big).
\end{equation}
This prediction is returned to the base frame and substituted for the measured
$^{B}T_{Tr}$ in Eq.~(\ref{eq:handtargets}),
\begin{equation}
\label{eq:gppushback}
^{B}\hat{T}_{Tr}(k{+}1) = {}^{B}T_{Tr}(0)\; ^{0}\hat{T}_{Tr}(k{+}1),
\end{equation}
so the hands are servoed to where the torso is predicted to be rather than
where it was last observed. No other term in Eq.~(\ref{eq:qp}) changes.

\subsection{Testing Phantom}

\subsubsection{Hardware}

For realistic testing without risking human subjects, we created a phantom with sensing capabilities and appropriate degrees of freedom to serve as the patient. The device can be seen in Fig \ref{fig: testing_man}. In the dissected view, the key components of the neck design for testing are visible, including a spherical bearing with infinite axial rotation and a cone limited to $45\degree$ to achieve $90\degree$ in the lateral left-right bending and flexion-extension directions. The base of the neck is then mounted to a transducer force-torque sensor, BOTA SensONE Gen A, which measures the translational forces applied to the neck.

\subsubsection{Software}
\label{sec: man_soft}

Tracking follows the same protocol as on the robotic system, but uses a set of OptiTrack cameras, the Trio V120. A series of markers were placed on both the head and torso. The neutral position, with the head and torso aligned, was defined with the basis aligned. As with the ArUco markers, the frames were aligned with the biomechanical frames to properly interpret motions relative to clinically relevant measures and limits. To accurately track forces, the BOTA sensor was similarly aligned with the biomechanical frame and contains an internal IMU, which removed the necessity to track gravity.

\begin{figure}
\vspace{-2mm}
    \centering
    \includegraphics[trim=22cm 22cm 20cm 18cm, clip, width=0.99\linewidth]{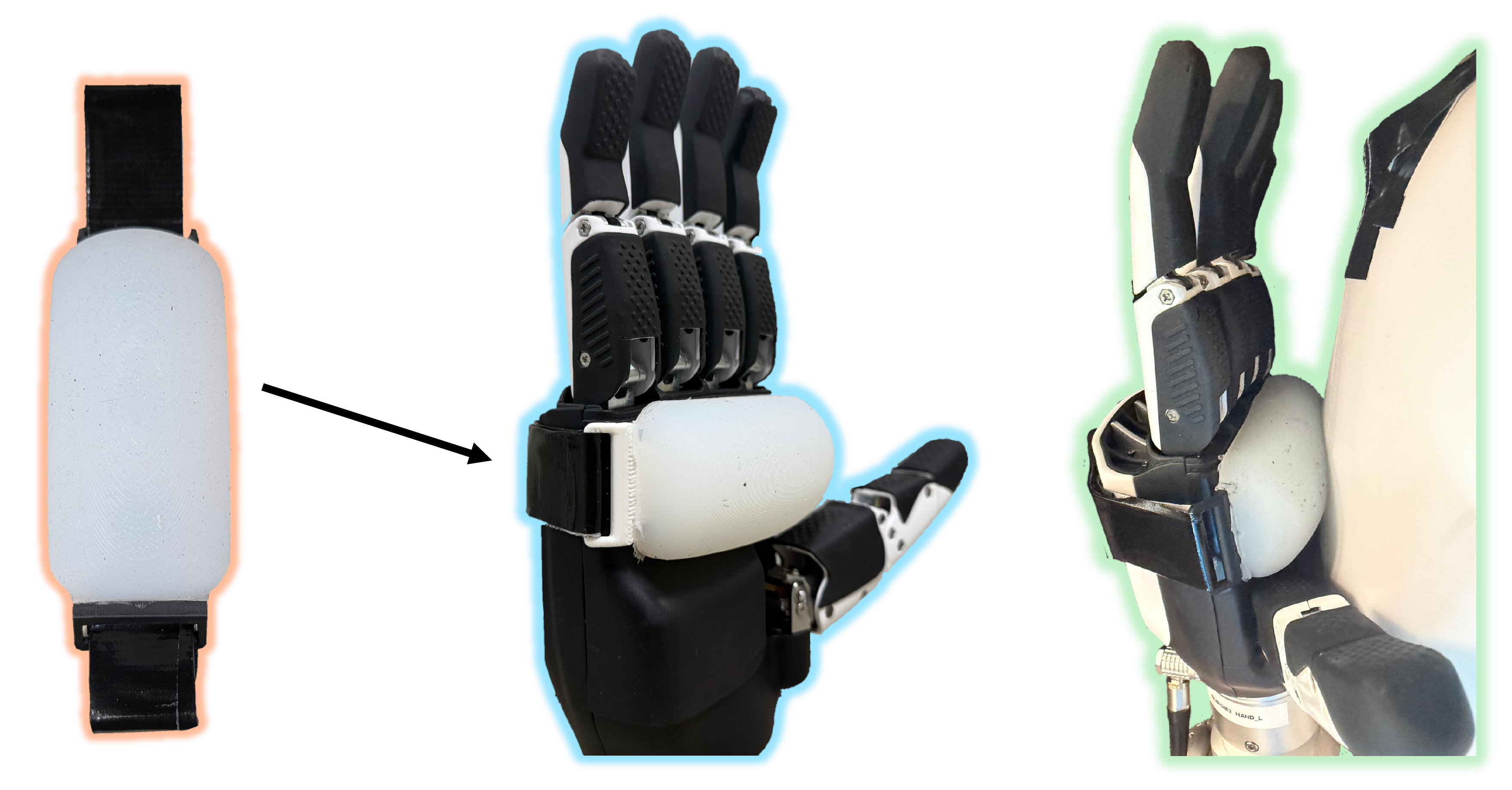}
    \caption{Flexible pad used to make hand softer and more compliant}
    \label{fig: hand_pad}
    \vspace{-7mm}
\end{figure}

\section{Experiments and Results}

\begin{figure*}
    \centering
    \vspace{2mm}
    \includegraphics[trim=2cm 34cm 2cm 8cm, clip, width=0.99\linewidth]{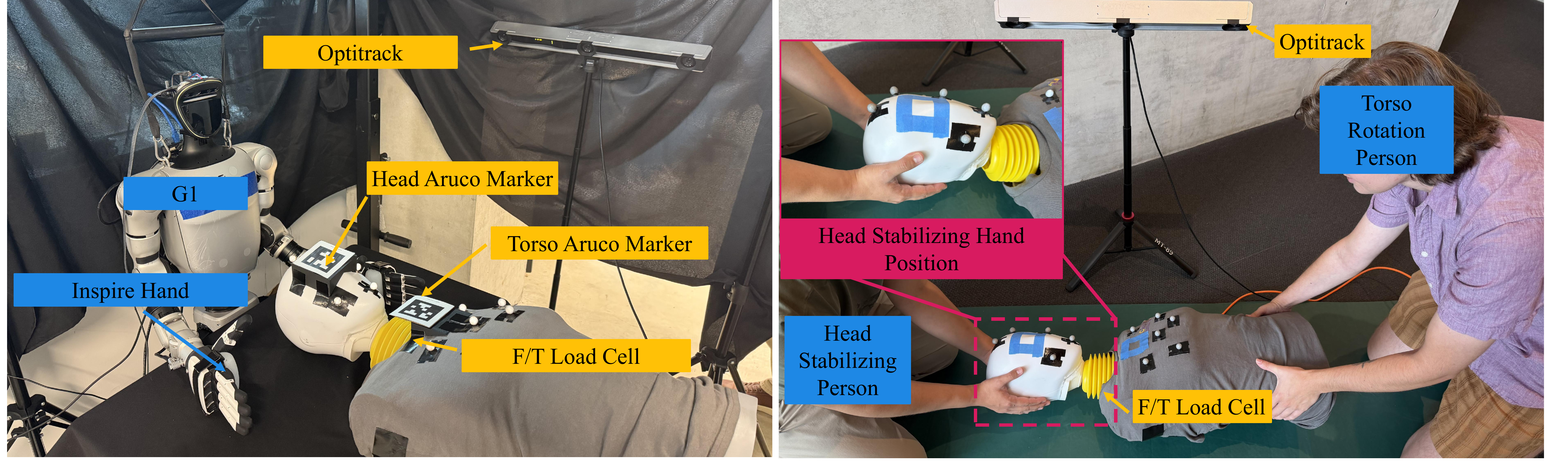}
    \caption{Hardware setup for experiments of head stabilization by a human (left) and by the robot (right).}
    \label{fig: test_set_up}
    \vspace{-6mm}
\end{figure*}

We placed heavy emphasis on safety metrics to evaluate the algorithm's performance and the platform's ability to execute this task. To compare robotic execution with human capabilities, we compared the robot's safety metrics and responses with human experience, including novices and top performers, or "experts".

\subsection{Experimental Hardware}

The tasks require hardware capable of tracking a head and torso, grasping a head, and rotating it. The hardware chosen for this task was the Unitree G1 humanoid, equipped with a D435 RealSense camera and a pair of Inspire RH56DFTP dexterous hands. The G1 has 7 degrees of freedom in each arm (14 for bi-manual tasks) and a head-mounted camera mounting point. 

Attached to the palms are semi-flexible pads that improve grip, as the Inspire hands' shape does not allow for a large grip surface relative to the round shape of the head. The pads are made from silicone and a 3D-printed skeleton with strap loops on either side, curved to minimize interference with hand or finger motion, allowing quick placement and removal. Figure \ref{fig: hand_pad} shows the pad on its own, on the hand, and actively pressing as it would during the task.

\subsection{Locating and Moving to Head}

\begin{figure}
    \centering
    \includegraphics[trim=25cm 28cm 20cm 8cm, clip, width=0.99\linewidth]{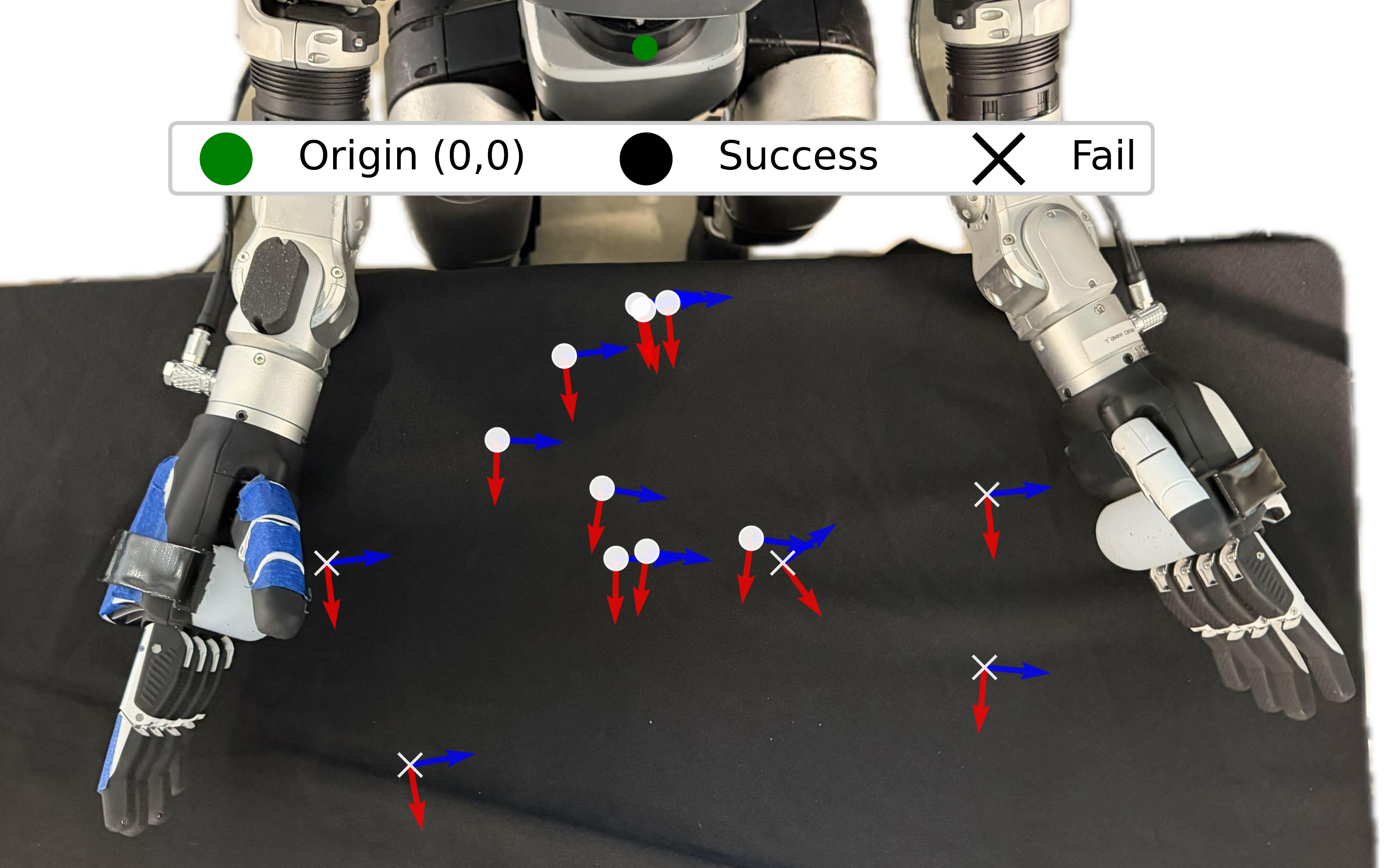}
    \caption{The head was posed in multiple locations across the workspace, and the results of finding a bi-manual set of trajectories were recorded. The "$\bullet$" indicates a successful ability to grasp, and the "x" represents a failure. The red and blue vectors show the orientation of each pose.}
    \vspace{-6mm}
    \label{fig: to_head_test}
\end{figure}

This test isolated the first step in the overall task, which requires the robot to locate the head and torso, then place its hands appropriately prior to making contact with the head. Across 14 trials with variations in the initial head pose, the robot successfully grasped 9 which were concentrated in front of the robot with a minimal allowed head angle. Figure \ref{fig: to_head_test} shows the pose variations. Orientation variation for successful lifts was limited because the G1's workspace is constrained by its smaller limb length and DOF relative to the human-sized phantom.

\subsection{Grasp Results}

Basic testing was performed by grasping the head with 3 different durometer silicone rubbers. The silicone rubber grips were placed in the robot's hands, as shown in Figure \ref{fig: hand_pad}. The robot was instructed to locate and grasp the head with a grip wrench force command. The BOTA sensONE Gen A F/T sensor measured the tugging force on the head to determine grip strength. These values, $f_y$ and $f_r$ taken from the experiment, were then used in equation \ref{eq: material_opt} to compare material choices. The Poisson ratio $\nu$ was assumed to be .5 for all pads. The Young's modulus for the Ecoflex 00-10 and 00-20 is 8 psi \cite{smoothon_ecoflex_tb} and the Dragon Skin 20 is 49 psi \cite{smoothon_dragonskin_tb}. 

Adding deformable, tacky inserts to the palm of the hand significantly improved successful grasping of the phantom head. Results in Table \ref{tab:force_pressure_ratios} show that the best performance was achieved using the Ecoflex silicone at 3-5 N of grasp force however, the Ecoflex 00-10 was too soft to be durable enough for testing, leading to the decision to use the Ecoflex 00-20 at 3N of grasp force, which provided a balance of high grip and low pressure applied to the skull.

\begin{table}[t]
\centering
\caption{Force--pressure ratios by material}
\label{tab:force_pressure_ratios}
\begin{tabular}{l c c c c}
\toprule
Material & $f_{y}$ (N) & Pressure (kPa) & $f_{r}$ (N) & Ratio \\
\midrule
\multirow{3}{*}{Ecoflex 00-10} & 3.0 & 12.43 & 19.57 & 1.574 \\
 & 5.0 & 14.74 & 15.45 & 1.048 \\
 & 7.0 & 16.49 & 14.27 & 0.865 \\
\midrule
\multirow{3}{*}{Ecoflex 00-20} & 3.0 & 12.43 & 18.72 & 1.506 \\
 & 5.0 & 14.74 & 18.83 & 1.277 \\
 & 7.0 & 16.49 & 19.86 & 1.204 \\
\midrule
\multirow{3}{*}{Dragon Skin 20} & 3.0 & 41.62 & 8.35 & 0.201 \\
 & 5.0 & 49.35 & 8.76 & 0.177 \\
 & 7.0 & 55.21 & 11.37 & 0.206 \\
\midrule
\multirow{3}{*}{None} & 3.0 & - & 8.05 & - \\
 & 5.0 & - & 11.75 & - \\
 & 7.0 & - & 7.53 & - \\
\bottomrule
\end{tabular}
\vspace{-6mm}
\end{table}

\subsection{Robotic Bi-manual Stability Performance}
\begin{figure*}[t]
    \centering
    \vspace{2mm}
    \setlength{\tabcolsep}{2pt}
    \begin{tabular}{>{\centering\arraybackslash}m{1.5em} *{5}{m{0.19\textwidth}}}
        \textbf{(A)} &
        \includegraphics[trim= 40mm 60mm 80mm 140mm, clip, width=0.19\textwidth]{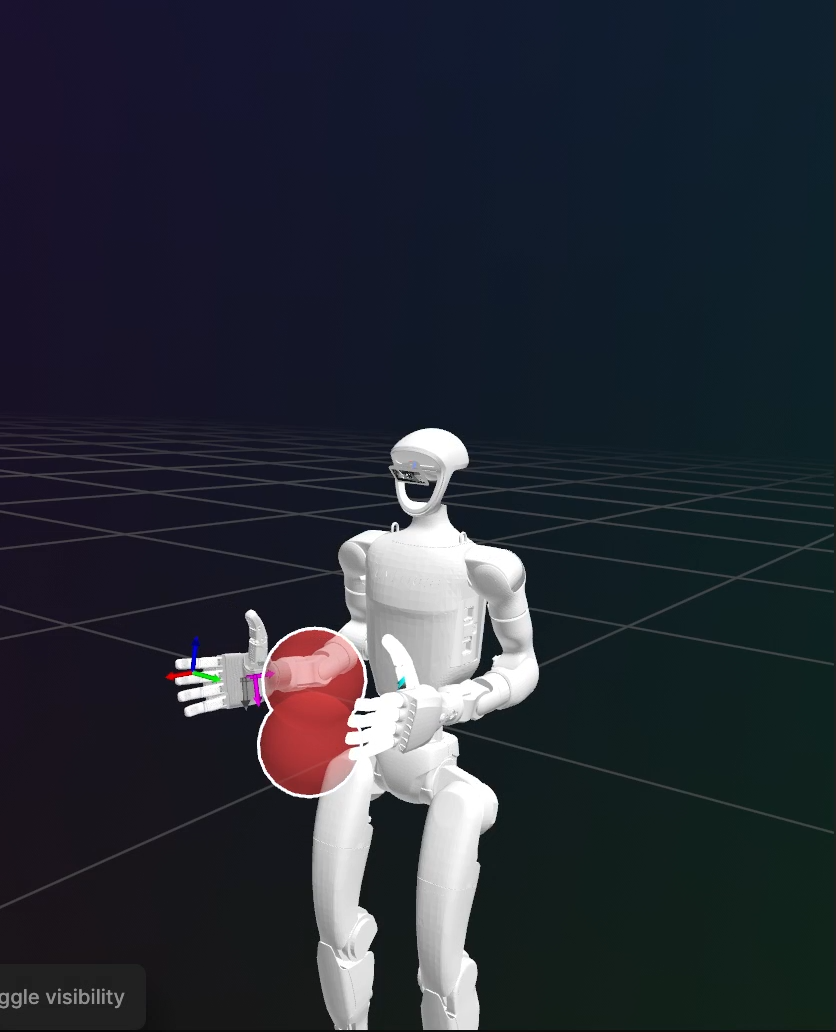} &
        \includegraphics[trim= 40mm 60mm 80mm 140mm, clip, width=0.19\textwidth]{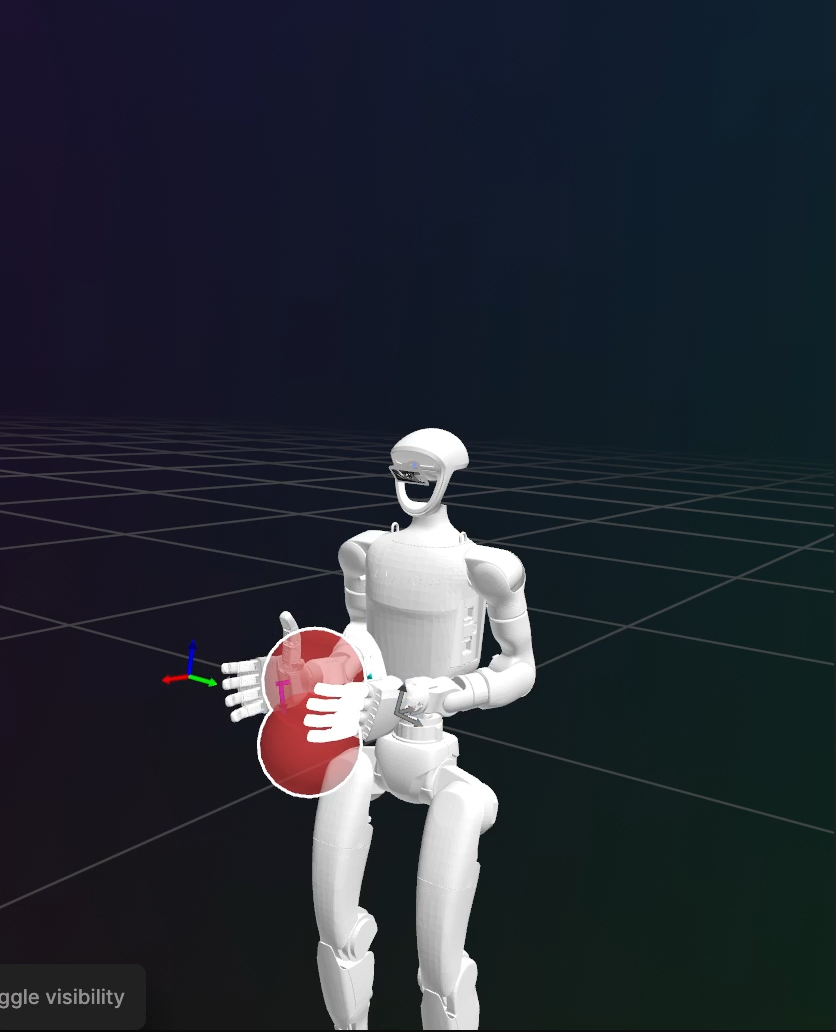} &
        \includegraphics[trim= 40mm 60mm 80mm 140mm, clip, width=0.19\textwidth]{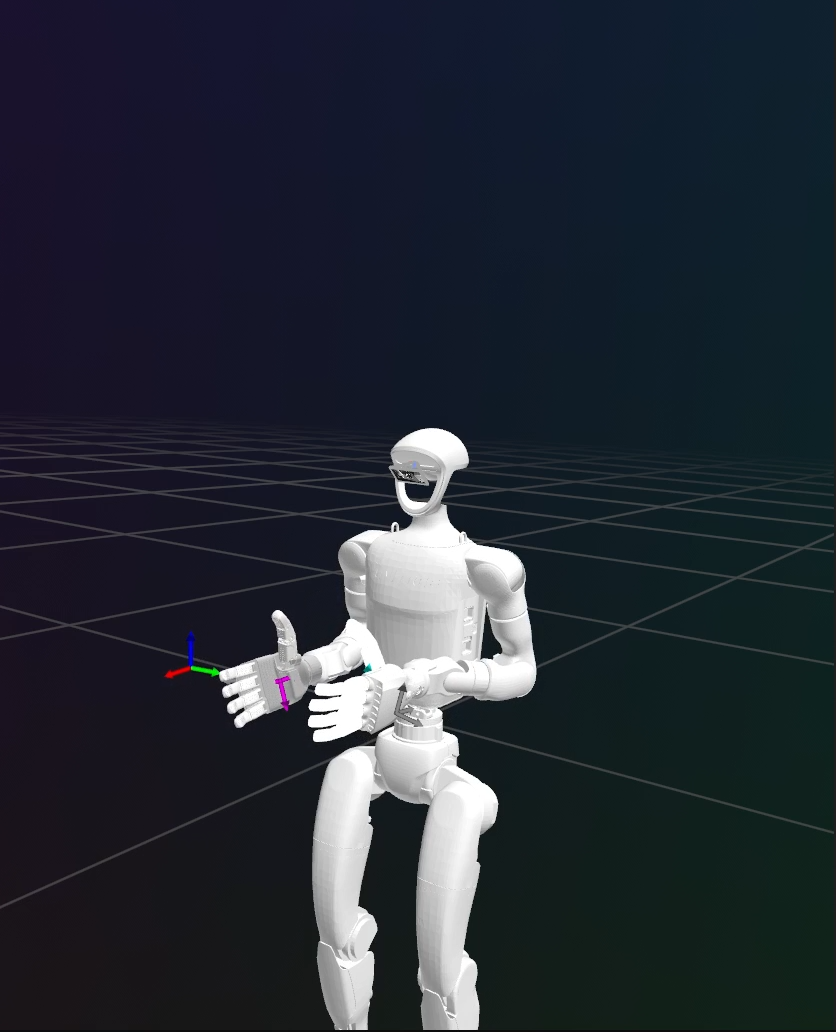} &
        \includegraphics[trim= 40mm 60mm 80mm 140mm, clip, width=0.19\textwidth]{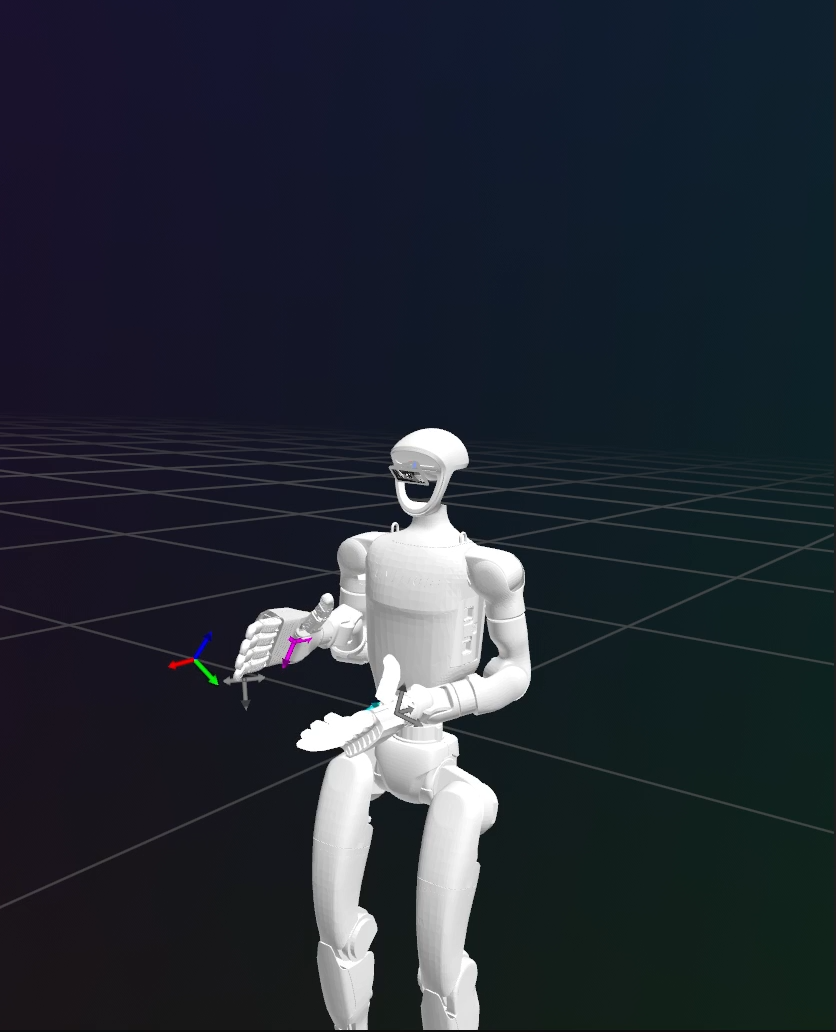} &
        \includegraphics[trim= 40mm 60mm 80mm 140mm, clip, width=0.19\textwidth]{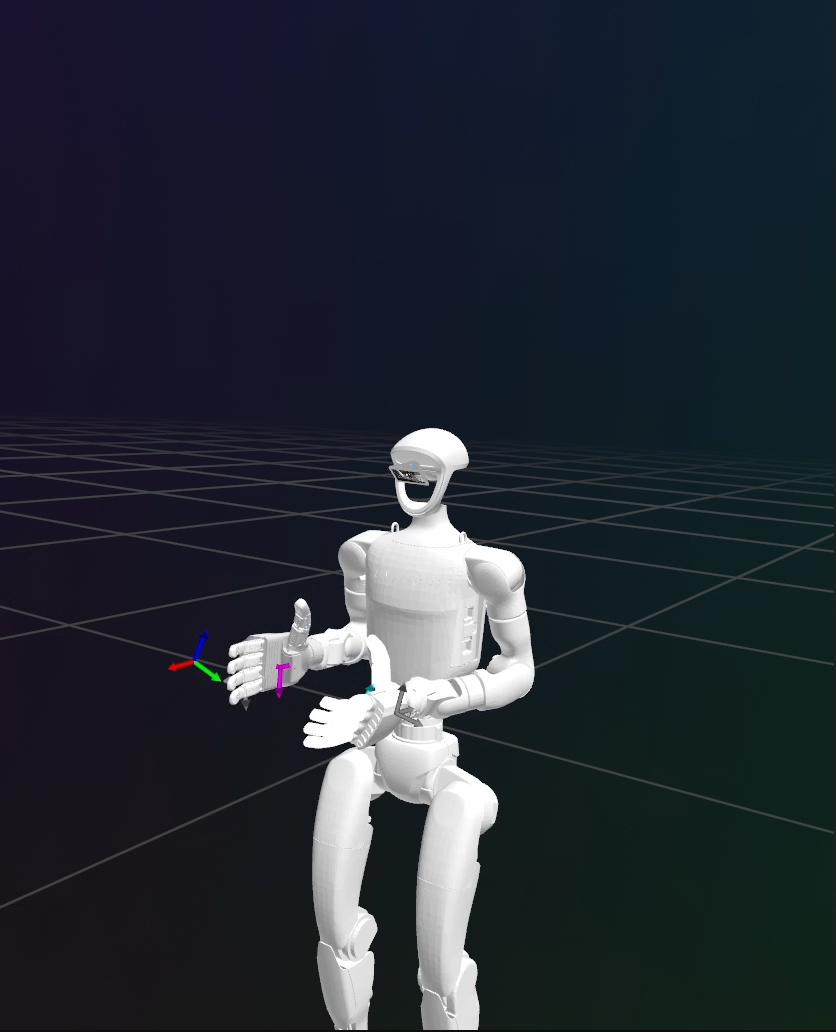} \\[1mm]

        \textbf{(B)} &
        \includegraphics[width=0.19\textwidth]{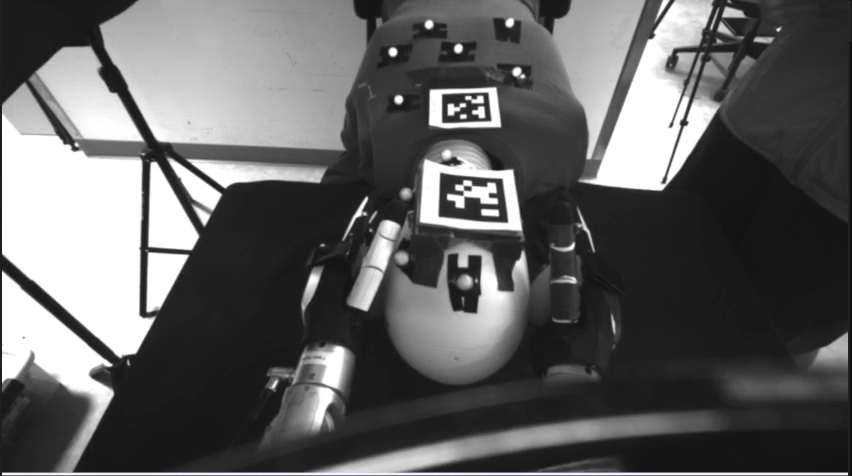} &
        \includegraphics[width=0.19\textwidth]{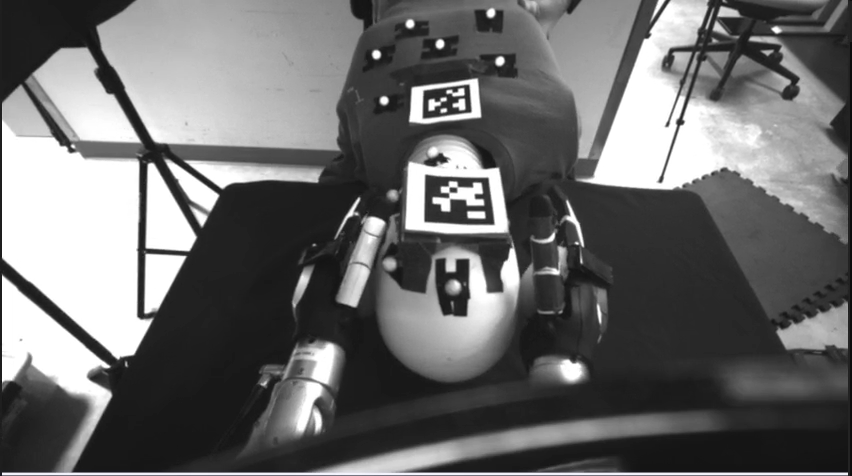} &
        \includegraphics[width=0.19\textwidth]{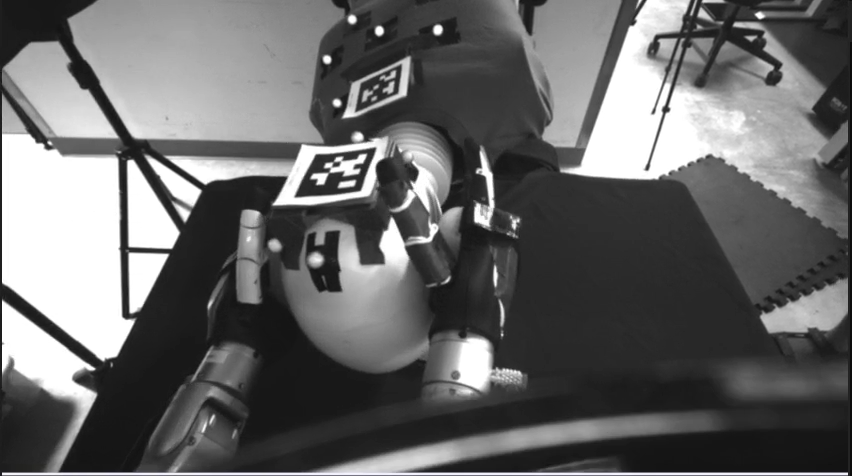} &
        \includegraphics[width=0.19\textwidth]{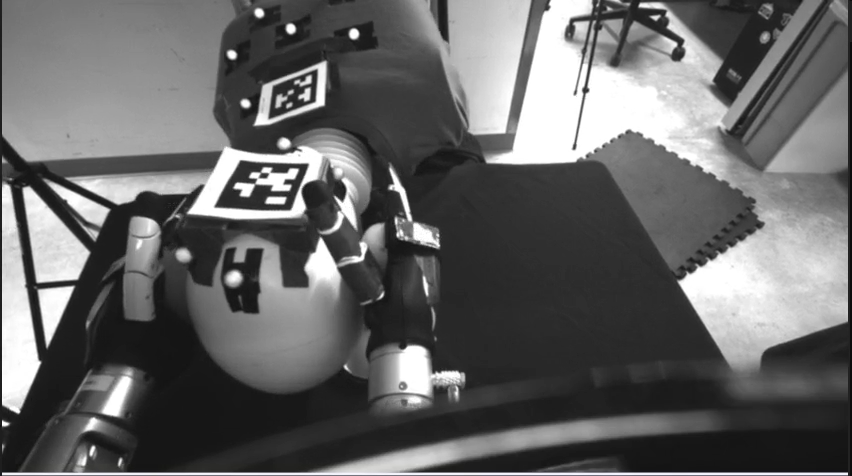} &
        \includegraphics[width=0.19\textwidth]{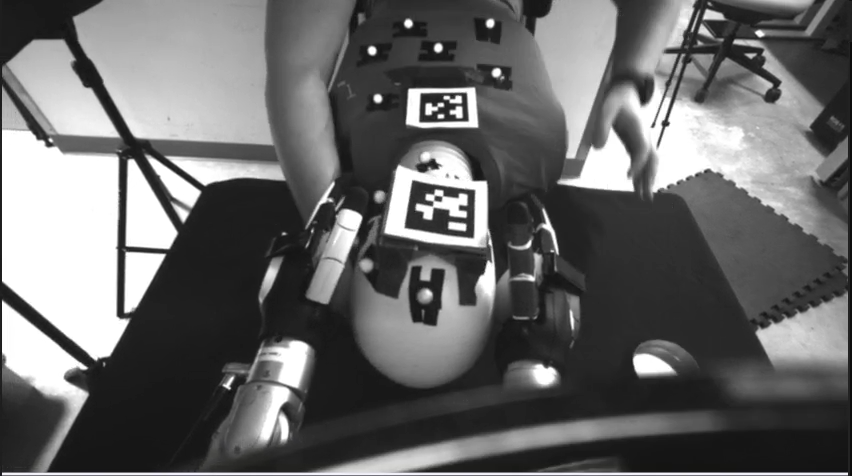} \\[1mm]

        \textbf{(C)} &
        \includegraphics[trim= 100mm 0 140mm 0, clip, width=0.19\textwidth]{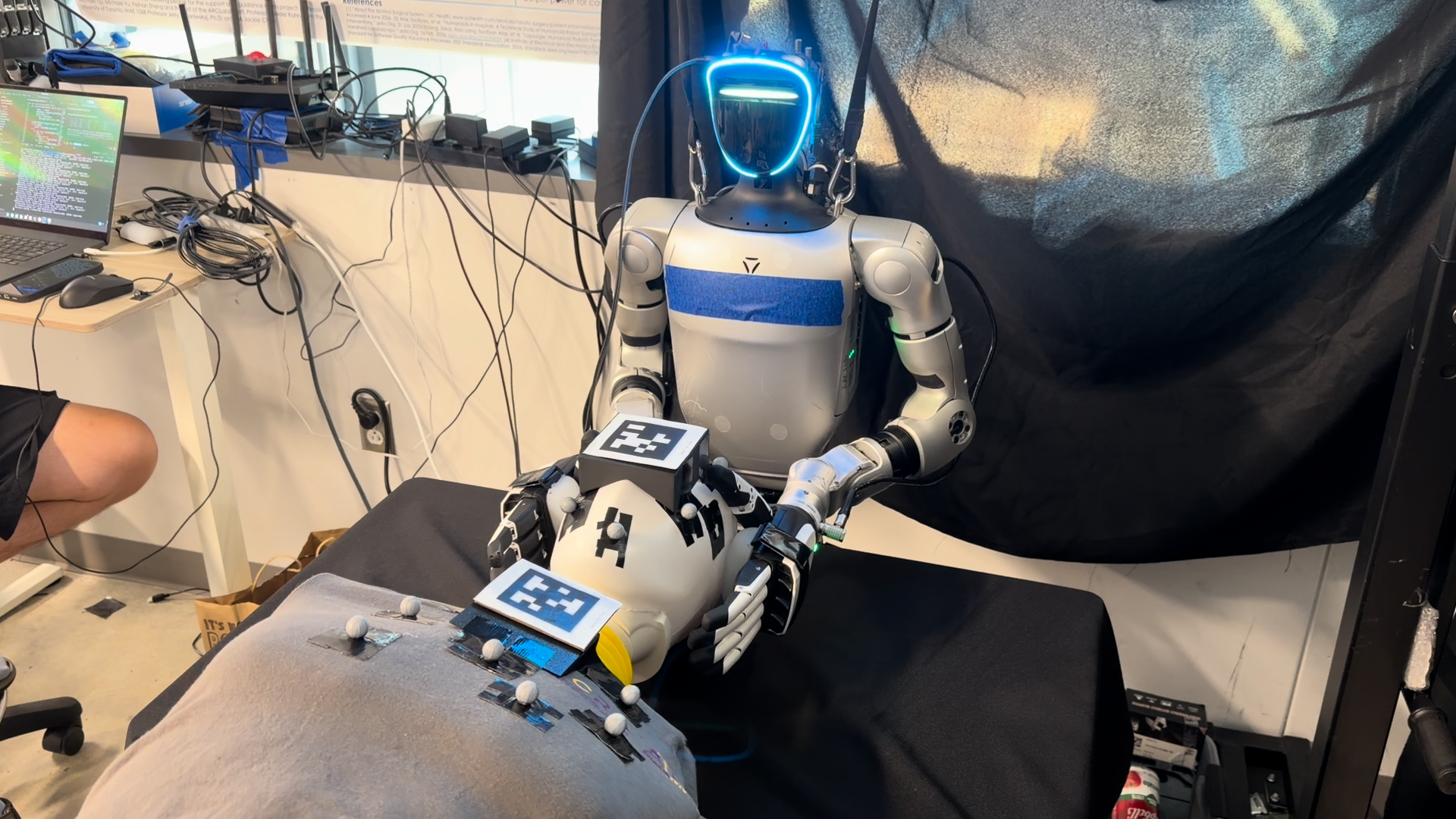} &
        \includegraphics[trim= 100mm 0 140mm 0, clip, width=0.19\textwidth]{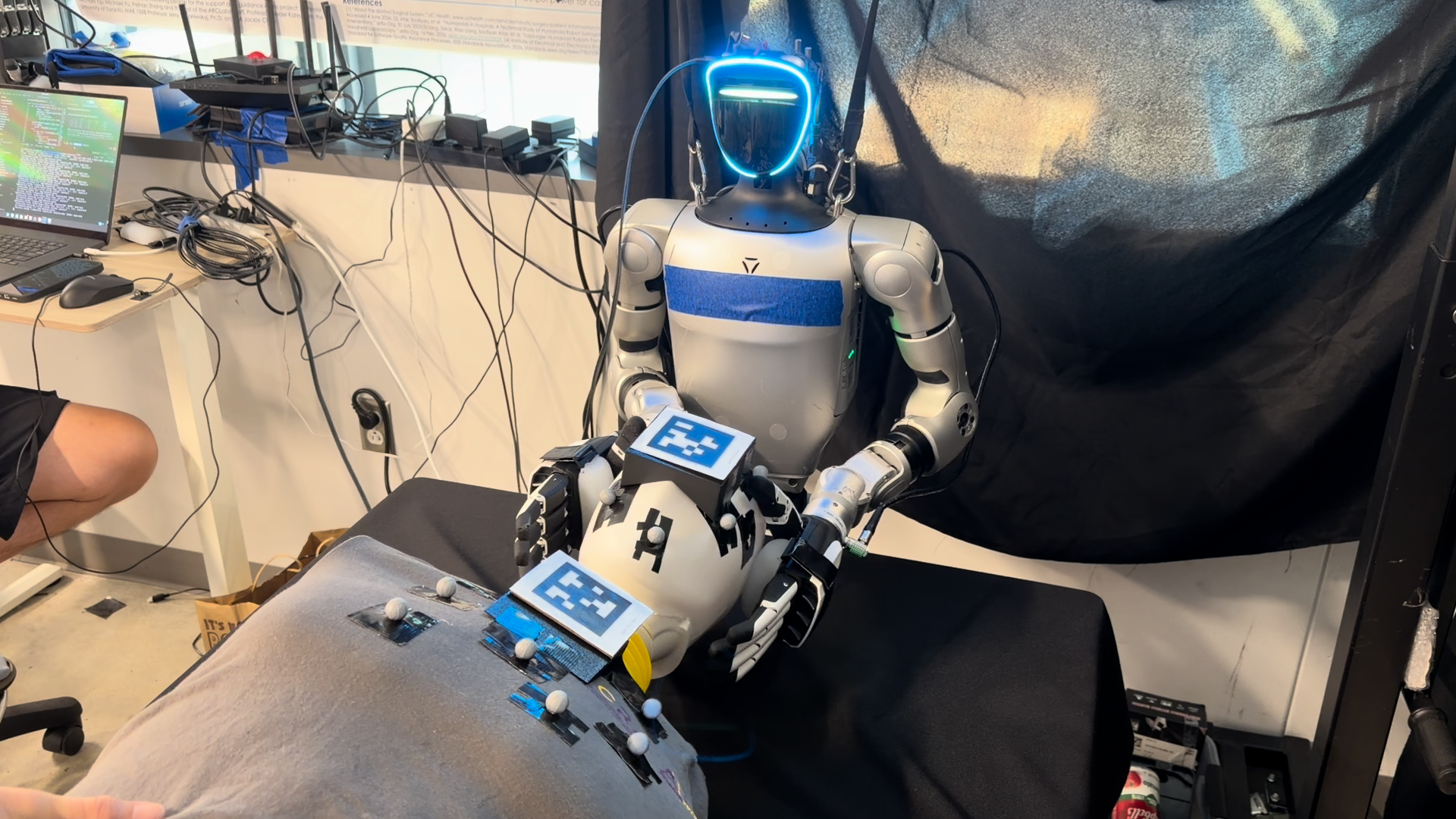} &
        \includegraphics[trim= 100mm 0 140mm 0, clip, width=0.19\textwidth]{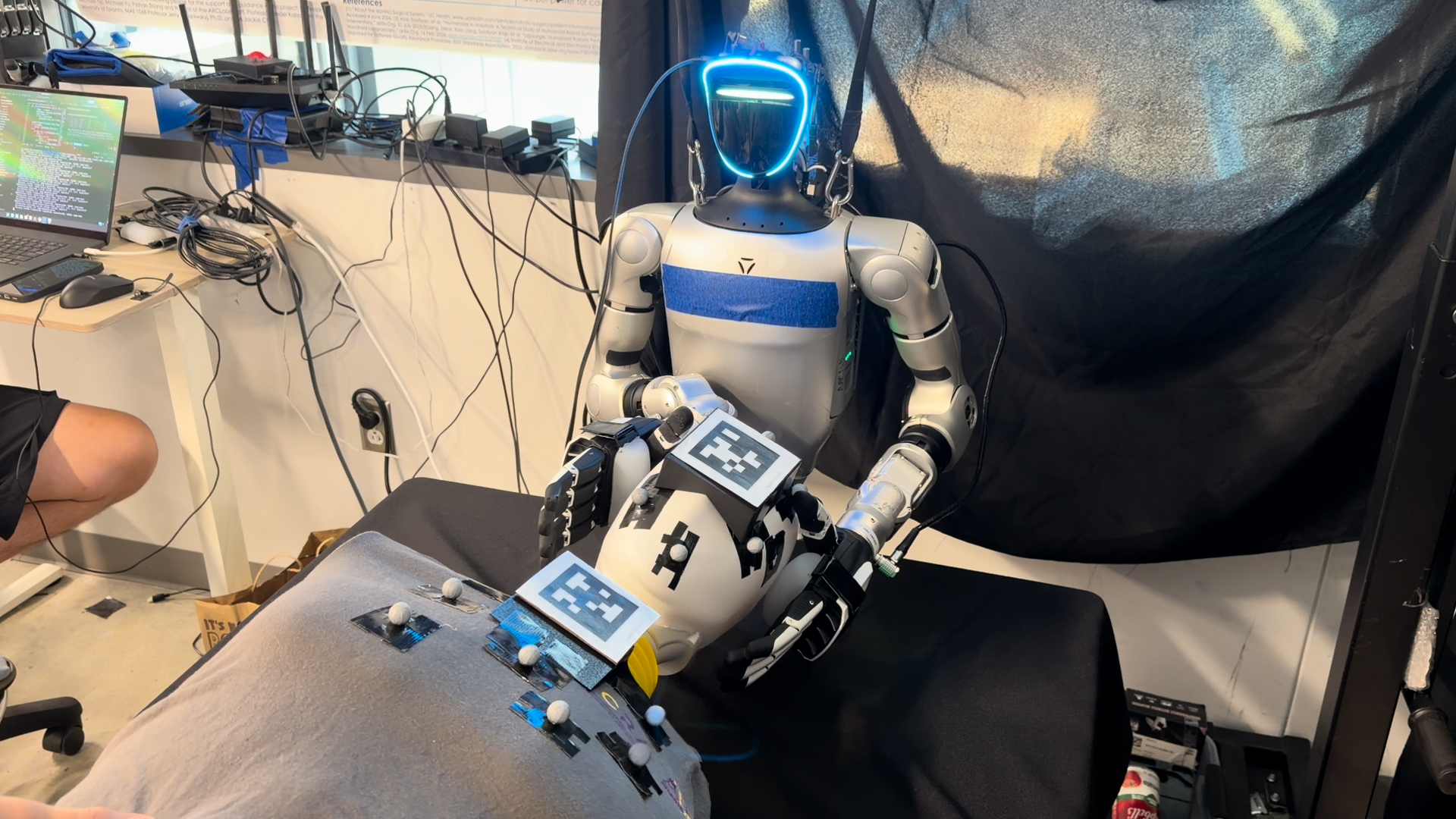} &
        \includegraphics[trim= 100mm 0 140mm 0, clip, width=0.19\textwidth]{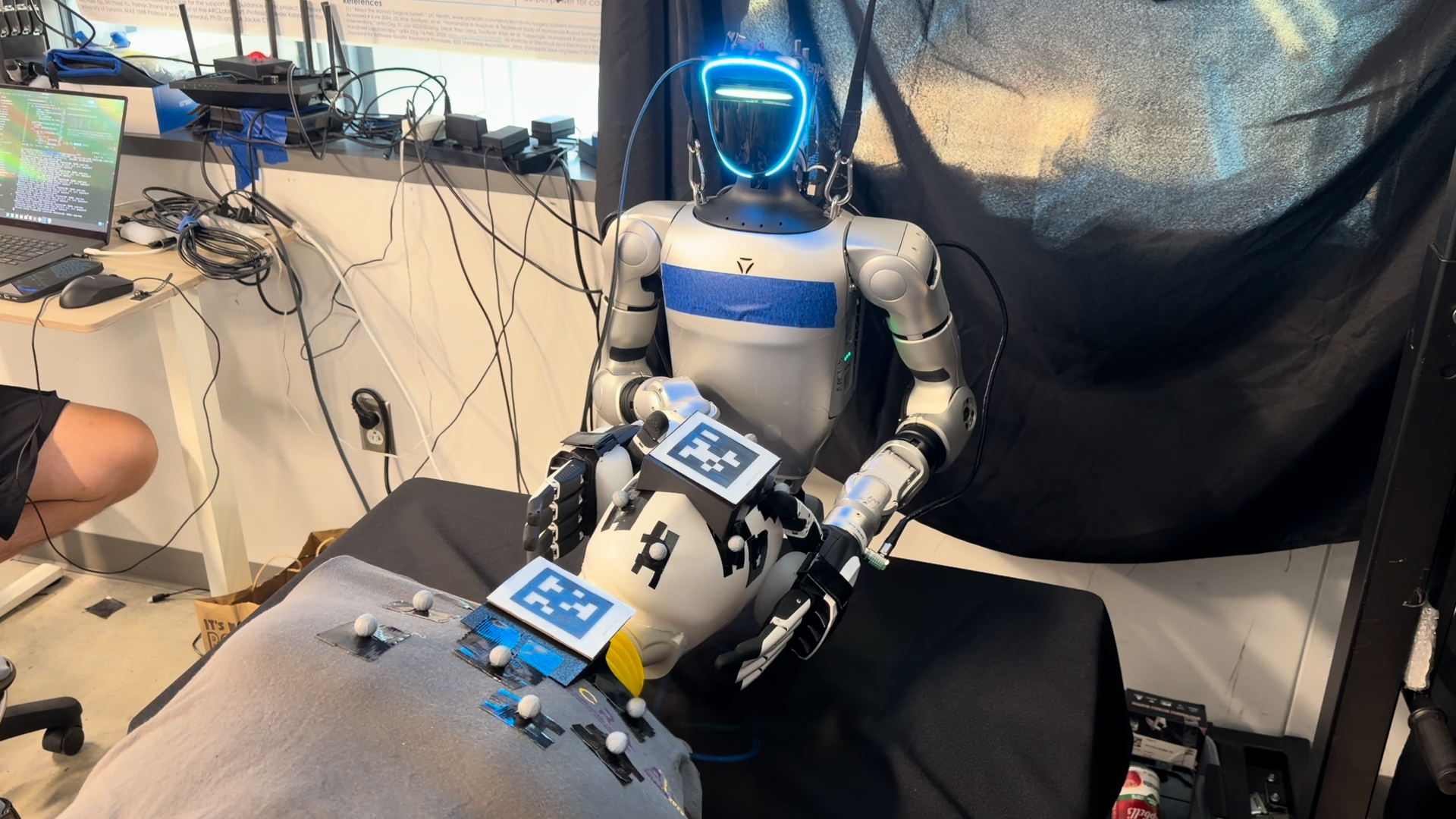} &
        \includegraphics[trim= 100mm 0 140mm 0, clip, width=0.19\textwidth]{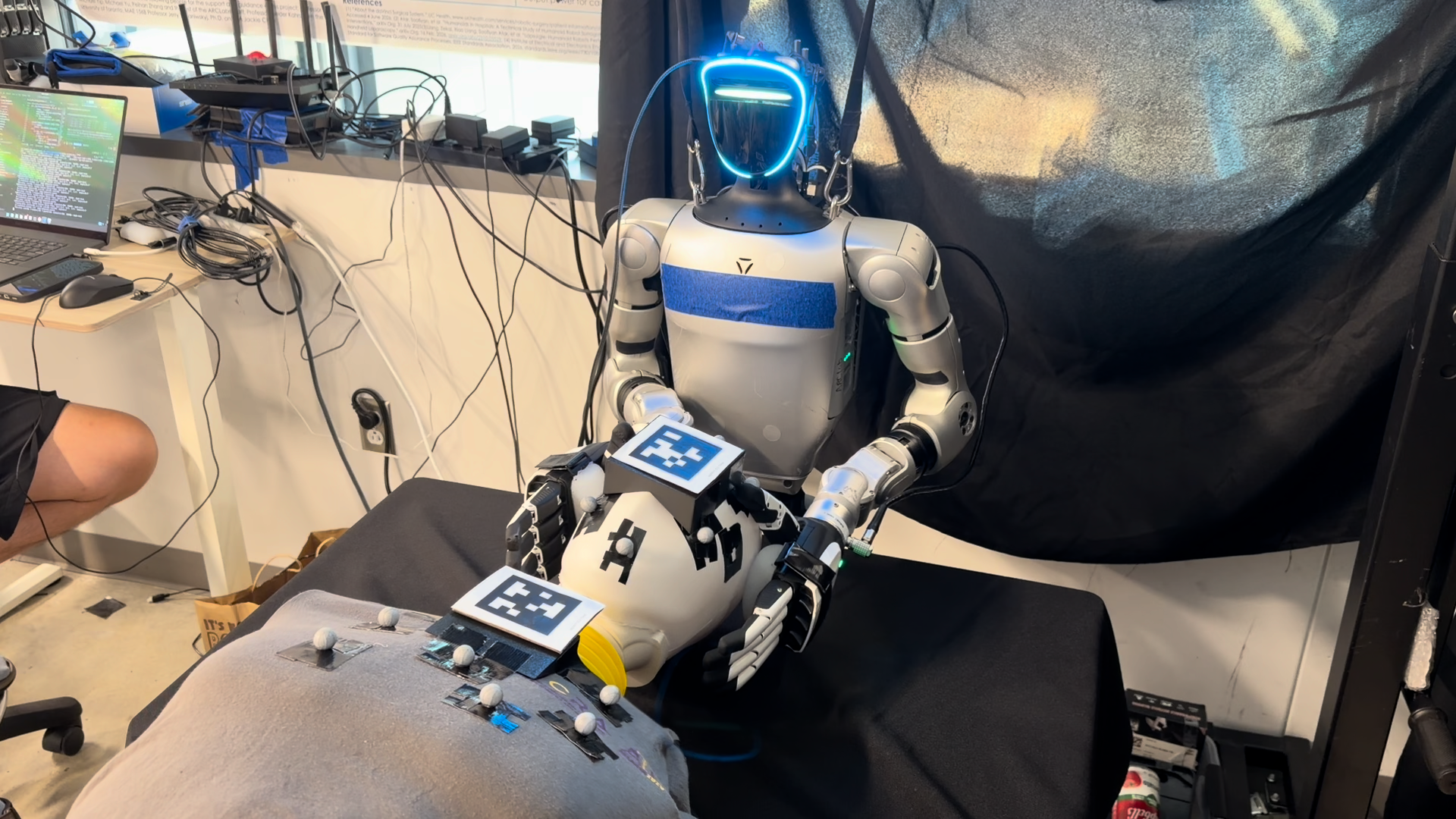} \\

    \end{tabular}
    \caption{(A) Simulated grasping and trajectory showing the head and floor collision boundaries. (B) Neck stabilization from the vantage point of the camera on the robot (C) Example neck-stabilizing sequence}
    \label{fig: seq}
    \vspace{-7mm}

\end{figure*}

\subsubsection{Protocol}

The robot protocol was conducted as described above; however, for the human, a slight variation in data collection was needed to gather the same dataset as was collected by the robotic system. 

\begin{figure}[t]
\vspace{2mm}
    \centering
    \includegraphics[width=.99\linewidth]{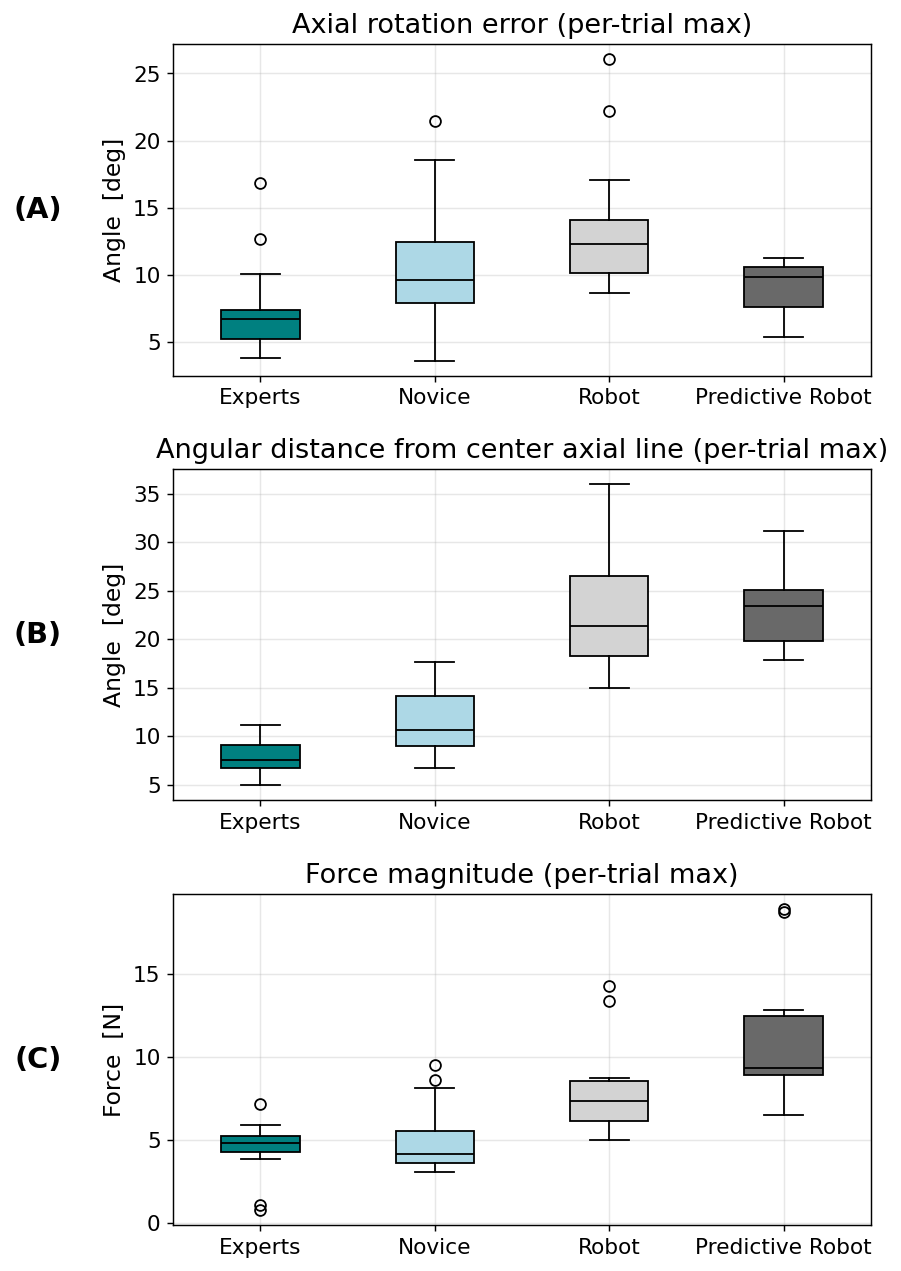}
    \caption{Comparison of expert (n=21) and novice (n=56) humans to a humanoid robot with (n=12) and without (n=12) a predictive algorithm in a neck stabilization task. (A) Maximum axial rotation error of head relative to torso. (B) Max angular distance error of neck bending relative to center axial line of torso. (C) Max force exerted on neck during neck stabilization task.}
    \label{fig:box_plots}
    \vspace{-8mm}
\end{figure}

The human protocol uses OptiTrack, the Trio V120 camera, on a tripod placed to the right or left of the phantom, as described in \ref{sec: man_soft} to track the head and torso poses relative to each other. The "user" is placed at the head and told to grasp the head with thumbs and fingers around where the ears would be. The "teammate" who rolls the torso is placed at the end of the torso on the side opposite the camera, as shown in Figure \ref{fig: test_set_up}. Once data collection begins, the "teammate" is instructed to roll the torso at a rate similar to the speed they would roll a human, and the "user" is told to maintain head alignment during the roll. "Users" were given a visual instruction on undesirable angles before their trials. Each roll is considered an individual trial of data collected to ensure accurate force readings for each trial. The results of this testing are shown in Fig \ref{fig:box_plots}.

\subsubsection{Human Performance}
The expert human group had a median maximum axial rotation error of 6.68\degree\ and the novice group had an error of 9.64\degree. The maximum angular distance from center axial line or combined bending and flexion/extension of the neck had a median of 7.59\degree\ for the experts and 10.59\degree\ for the novices. 

\subsubsection{Robotic Performance}

The robot was tasked with locating and grasping the head, then following the torso rotation while keeping the head aligned. The robot with no prediction model kept alignment within 13.66\degree\ in the axial direction, within 23.12\degree\ for the other degrees of freedom, and maintained a force under 8N. The robot with the prediction model had a median maximum axial rotation of 9.85\degree\ and 23.4\degree\ in the other directions. The digital twin, robot camera view and external view of an executed roll shown in Fig \ref{fig: seq}.

While overall robot performance did not match human capabilities, some trials met the requirements and achieved metrics comparable to human performance in this test. The test was very challenging for this particular robot, as the constrained bi-manual task reduces the degrees of freedom available for null-space control and optimization. For this particular system, the null space goes from 2 DOF, one for each arm, to 1 DOF due to the extra constraint added by the grasp. The G1 is also fairly small relative to the human-sized phantom, making the head-to-workspace ratio quite small. The prediction model, while better at alignment, did not react to disturbances as well and thus produced higher reaction forces along the neck base.

\vspace{-2mm}
\section{Discussion and Conclusion}
This paper presented an algorithm for head stabilization and a method to test and compare it against novice humans on a level playing field not seen in other works of it's kind. The robot with a predictive model consistently produced axial rotational alignment within the novice human's range, with a median below the maximum allowable deviation. While the robot's abilities are not yet at human-level performance for preventing neck-bending stabilization, future, more human-like high-DOF hardware may deliver higher performance. With limited null space it become difficult to optimize for multiple head angles on top of the reaction force minimization. The additional grippers significantly increased grasp strength without causing the robot to overheat from constantly regulating high force between the hands.  
One area of improvement would have been integration of high-fidelity force sensing at the robot's end effectors to either deploy a version of an impedance controller such that it doesn't interfere with the grip directions, which would again benefit from increased DOF, or to use the force sensing as feedback into the planner to pick more optimal trajectory points. 
\vspace{-2mm}
\section{Acknowledgments}
\vspace{-2mm}
Claude Code was utilized to help implement the described code/algorithms and scripts to graph the visualized data. Gemini generated the images in Figure 2.
\vspace{-2mm}

\balance
\bibliographystyle{ieeetr}
\bibliography{refs}
\balance

\end{document}